\pdfoutput=1
\documentclass[11pt]{article}
\usepackage[final]{acl}
\usepackage{times}
\usepackage{latexsym}
\usepackage[T1]{fontenc}
\usepackage[utf8]{inputenc}
\usepackage{microtype}
\usepackage{inconsolata}
\usepackage{graphicx}
\usepackage{amsmath}
\usepackage{amssymb}
\usepackage{mathtools}
\usepackage{booktabs}
\usepackage{multicol}
\usepackage{multirow}
\usepackage{colortbl}
\usepackage{xcolor}
\usepackage{algorithm}
\usepackage{algpseudocode}
\usepackage{array}
\usepackage{subcaption}
\usepackage{tabularx}
\usepackage{makecell}
\usepackage{placeins}
\usepackage{float}
\usepackage{lscape}
\usepackage{amsthm}
\usepackage{enumitem}
\usepackage{pifont}
\usepackage{listings}
\usepackage{fontawesome}
\usepackage{wrapfig}
\usepackage{caption}
\usepackage{mdframed}
\usepackage[most]{tcolorbox}

\title{Taming Extreme Tokens: Covariance-Aware GRPO with Gaussian-Kernel Advantage Reweighting}

\author{%
  Cheng Wang$^\dagger$ ~~ Qin Liu$^\ddagger$ ~~ Wenxuan Zhou$^\mathsection$ ~~ Muhao Chen$^\ddagger$\\[2pt]
  $^\dagger$National University of Singapore \quad
  $^\ddagger$University of California, Davis \\ $^\mathsection$University of Southern California \\
  \texttt{wangcheng@u.nus.edu}
}

\begin{document}
\maketitle

\begin{abstract}
Group Relative Policy Optimization (GRPO) has 
emerged as a promising approach for
improving the reasoning capabilities of large language models. However, 
it struggles to effectively balance the tradeoff between exploration and exploitation during training, 
often resulting in suboptimal performance.
Motivated by the theoretical 
insight that changes in entropy are governed by the covariance between token probabilities and their corresponding advantages, we propose a hyperparameter-free, \emph{covariance-weighted optimization} method that dynamically down-weights extreme token-level updates via a Gaussian kernel. This approach automatically 
reduces the instability caused by exploration-exploitation trade-off while preserving informative learning signals. 
Extensive empirical evaluations show that our approach 
improves downstream performance across reasoning benchmarks compared with GRPO, and effectively 
stablizes entropy as training progresses.
\end{abstract}

\section{Introduction}
Group Relative Policy Optimization (GRPO)~\citep{grpo} 
has emerged as a promising approach for 
enhancing the reasoning capabilities of large language models (LLMs), particularly in complex mathematical and coding tasks. Despite its demonstrated effectiveness, GRPO faces a 
critical limitation in properly
balancing between exploitation and exploration during policy optimization, which can undermine its performance~\citep{wang2025eframedeeperreasoningexplorationfilterreplay,wang2025eframedeeperreasoningexplorationfilterreplay}.

Excessive exploitation can cause the model to become overconfident in its suboptimal solutions, thereby limiting its capabilities to explore novel reasoning strategies and potentially overlook more effective approaches. Conversely, while exploration is necessary for identifying better policies, excessive exploration 
may result in unstable training dynamics and hinder convergence to 
a stable, high-performing solution. 
These opposing risks highlights the importance of a principled mechanism for balancing exploration and exploitation so as to realize a more robust GRPO.


\begin{figure}[!t]
    \centering
    \includegraphics[width=0.9\linewidth]{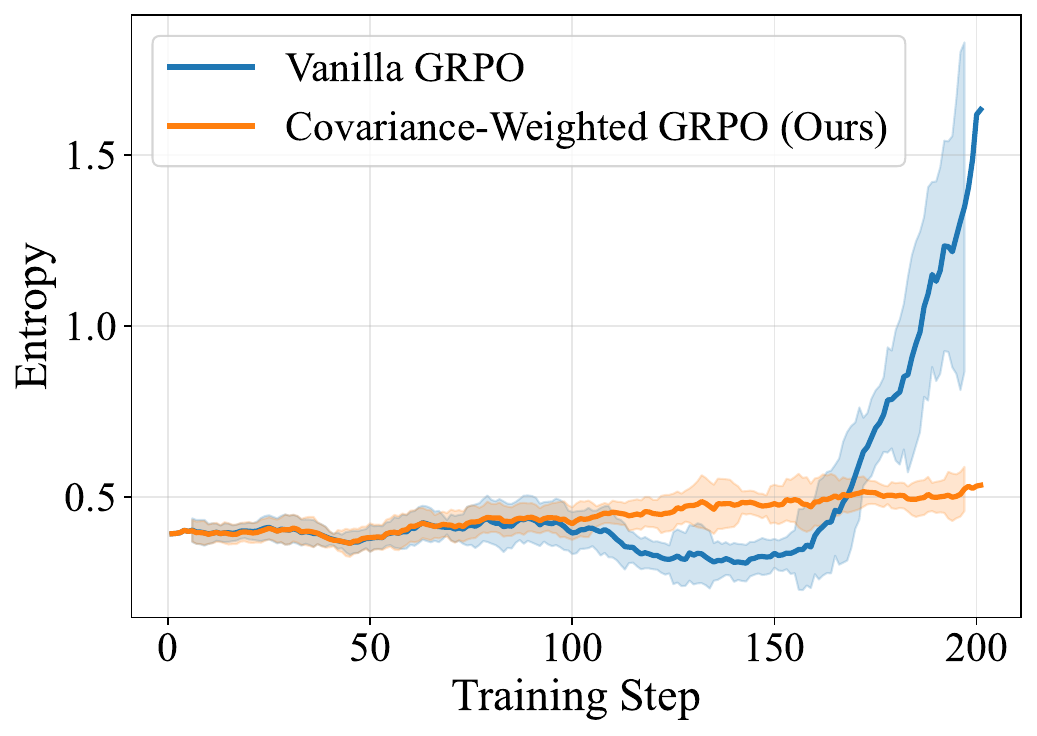}
    \caption{\textbf{Policy Entropy During Training.} Vanilla GRPO exhibits entropy instability, while our method keeps entropy at a reasonable level that effectively balances exploration and exploitation.}
    \label{fig:entropy}
\end{figure}

\begin{figure*}
    \centering
    \includegraphics[width=0.76\linewidth]{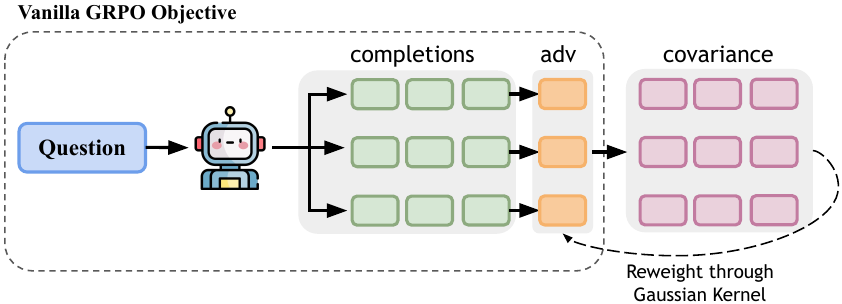}
    \caption{\textbf{Illustration of Our Proposed Method.} Compared with vanilla GRPO, our method reweights the advantages based on the covariance between token probabilities and advantages.}
    \label{fig:illustration}
\end{figure*}

Specifically,
the trade-off in GRPO is fundamentally tied to the policy's entropy dynamics during training. As established by \citet{entropymechanism}, entropy changes under the natural policy gradient update are governed by the covariance between token log-probabilities and their corresponding advantage estimates. Based on this theoretical foundation and our empirical observations, we identify that a small fraction of tokens with extreme covariance values disproportionately dominate the policy updates, resulting in entropy instability and degraded training dynamics.

To mitigate this issue, we propose a covariance-aware variant of GRPO that attenuates extreme token-level updates through Gaussian kernel weighting. Specifically, our approach computes the covariance between centered log-probabilities and centered advantages for each token, 
and applies a smooth down-weighting function to tokens with high-magnitude covariances while preserving the influence of those with moderate covariances. 
This mechanism effectively regulates the contribution of outlier tokens to the policy gradient, thereby improving the balance between exploration and exploitation in a hyperparameter-free way.
Extensive experiments demonstrate that our approach 
drastically improves over the vanilla GRPO, achieving better downstream performance and maintaining stable entropy dynamics, as illustrated in Figure~\ref{fig:entropy}.

\section{Method}

\subsection{Preliminaries}
GRPO~\citep{shao2024deepseekmath} extends Proximal Policy Optimization ~\citep{ppo} by removing the value network and using group-based rewards to estimate advantages. For each prompt $q$ sampled from $\mathcal{D}$, GRPO samples a group of $G$ responses $\{o_1, o_2, \ldots, o_G\}$ from the current policy $\pi_\theta$ and evaluates them using a reward model $r_\phi$, which is usually a rule-based verifier. GRPO computes the advantage for response $i$ as $\hat{A}_i = \frac{r_i - \bar{r}}{\sigma_r}$, where $r_i$ is the reward for response $i$, and $\bar{r}$ and $\sigma_r$ are the mean and standard deviation of rewards within the group. The GRPO aims to maximize the following objective:
\begin{align*}
J_{GRPO}(\theta) = \mathbb{E}_{q \sim \mathcal{D}} \left[ \frac{1}{G} \sum_{i=1}^G \frac{\pi_\theta(o_i|q)}{\pi_{\theta_{old}}(o_i|q)} \hat{A}_i \right] \\
- \beta \mathbb{E}_{q \sim \mathcal{D}} \left[ D_{KL}[\pi_\theta(\cdot|q) \| \pi_{ref}(\cdot|q)] \right],
\end{align*}
where $\pi_{\theta_{old}}$ is the policy from the previous iteration, $\pi_{ref}$ is the reference policy, and $\beta$ is the KL penalty coefficient.

\subsection{Motivation}
\label{sec:motivation}

To measure the exploration-exploitation trade-off in GRPO, we can use policy entropy as an indicator, which is defined as:

{
\begin{gather*}
\mathcal{H}(\pi_\theta) 
= -\mathbb{E}_{q \sim \mathcal{D}} \left[ \mathbb{E}_{o \sim \pi_\theta(\cdot|q)} [\log \pi_\theta(o|q)] \right] \\
= -\frac{1}{|\mathcal{D}|} \sum_{q \in \mathcal{D}} \frac{1}{|o|} \sum_{t=1}^{|o|} \mathbb{E}_{o_t \sim \pi_{\theta}} \left[ \log \pi_{\theta} \left( o_t \mid q, o_{<t} \right) \right].
\end{gather*}
}

\begin{figure}
    \centering
    \includegraphics[width=0.81\linewidth]{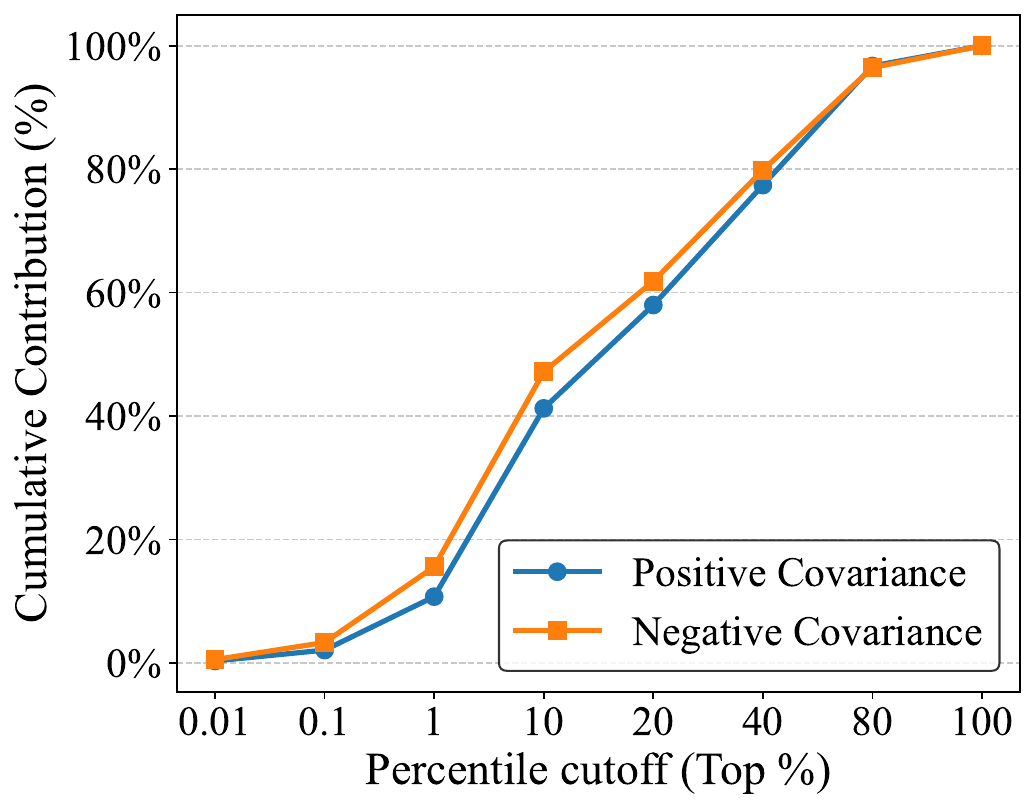}
    \caption{\textbf{Cumulative Contribution of Covariance Values}. A small fraction of tokens with extreme covariance values disproportionately dominate policy updates,}
    \label{fig:distribution}
\end{figure}

\citet{entropymechanism} show that, under Natural Policy Gradient~\cite{NIPS2001_4b86abe4}, the change in policy entropy is governed by the covariance between token log-probabilities and advantages:
\[
\Delta \mathcal{H} \approx -\eta \cdot \text{Cov}_{t}(\log \pi_{\theta}(o_t|q, o_{<t}), \hat{A}_i).
\]
This relationship reveals that the covariance between log-probabilities and advantages directly drives entropy dynamics during training. As a preliminary experiment, we use GRPO to fine-tune DeepSeek-R1-Distill-Qwen-1.5B~\cite{deepseekai2025deepseekr1incentivizingreasoningcapability} and track token-level covariance throughout 50 training steps. As shown in Figure~\ref{fig:distribution} and Table~\ref{tab:covariance_analysis}, we observe that a small fraction of tokens possess large-magnitude covariance values, which disproportionately dominate the overall covariance and precipitate unstable entropy dynamics. These extreme values push the policy away from ideal exploration-exploitation balance, leading to suboptimal performance. This observation directly leads to our method, which moderates extreme covariance updates through covariance-aware advantage reweighting, as suppressing these extreme updates is crucial for maintaining stable entropy.

\begin{table}[h]
\footnotesize
    \centering
    \resizebox{0.87\linewidth}{!}{%
    \begin{tabular}{l c c}
        \toprule
        \textbf{Percentile} & \textbf{Positive Covariance} & \textbf{Negative Covariance} \\
        \midrule
        0.01\% & 11.52 & -13.62 \\
        1.00\% & 3.32 & -3.34 \\
        20.00\% & 0.58 & -0.36 \\
        40.00\% & 0.33 & -0.22 \\
        100.00\% & 0.06 & -0.04 \\
        \bottomrule
    \end{tabular}%
    }
    \caption{\textbf{Covariance Distribution Statistics.} The table presents numerical values of covariance magnitudes at specific percentile thresholds.}
    \label{tab:covariance_analysis}
\end{table}

\subsection{Covariance-Aware Advantage Reweighting}
To address the issue of extreme covariance values destabilizing training so as to balance the exploration-exploration trade-off, we propose a covariance-weighted GRPO variant (CW-GRPO) that automatically down-weights tokens with large-magnitude covariances while preserving learning signals from moderate-covariance tokens.

We adopt the standard GRPO setup in which a policy $\pi_\theta$ produces $G$ responses $o_i$ per prompt $q$. The vanilla GRPO objective is
\begin{align*}
\mathcal{L}_{\text{GRPO}} &= \mathbb{E}_{q,\{o_i\}}
\Bigl[\frac{1}{G}\sum_{i=1}^G\frac{1}{|o_i|}
      \sum_{t=1}^{|o_i|}\ell_{i,t}\Bigr],
\end{align*}
where $\ell_{i,t}$ is the token-level loss defined as:
\begin{align*}
\ell_{i,t} = 
 \frac{\pi_\theta(o_{i,t}|q, o_{i,<t})}{\pi_{\theta_{old}}(o_{i,t}|q, o_{i,<t})} \hat{A}_i - \beta \text{KL}_{i,t}.
\end{align*}

Based on the motivation from Section~\ref{sec:motivation}, we propose to reweight the advantage signal based on the covariance between log-probabilities and advantages. We compute the token-level covariance between centered log-probabilities and centered advantages:
$$c_{i,t} = \left(\log \pi_\theta(o_{i,t}|q, o_{i,<t}) - \overline{\log \pi}\right) \cdot \left(\hat{A}_i - \overline{\hat{A}}\right),$$
where $\overline{\log \pi}$ and $\overline{\hat{A}}$ are the means of log-probabilities and advantages computed over responses.

We then apply a Gaussian kernel that exponentially suppresses only the magnitudes that exceed typical variability by setting the bandwidth to the empirical standard deviation $\sigma$ of the covariances:
\[
w_{i,t}=\exp{\Bigl(-\tfrac{c_{i,t}^2}{2\sigma^2}\Bigr)},
\]
where $\sigma$ is the standard deviation of $\{c_{i,t}\}$. The Gaussian kernel softly filters out the handful of extreme covariance tokens that would otherwise destabilize entropy, yet leaves informative moderate-covariance tokens intact.

To maintain the expected loss, we normalize the weights:
\[
\tilde{w}_{i,t}=w_{i,t}\cdot\frac{N}{\sum_{j=1}^G\sum_{k=1}^{|o_j|}w_{j,k}},
\]
where $N = \sum_{j=1}^G|o_j|$ is the total number of tokens across all responses.

The covariance-weighted advantage reweighting modifies the token-level loss as:
\begin{align*}
\tilde{\ell}_{i,t} = 
 \frac{\pi_\theta(o_{i,t}|q, o_{i,<t})}{\pi_{\theta_{old}}(o_{i,t}|q, o_{i,<t})} \cdot (\tilde{w}_{i,t} \hat{A}_i) - \beta \text{KL}_{i,t}.
\end{align*}
Our advantage reweighting approach automatically maintains the policy entropy at a reasonable level by adaptively filtering extreme covariance tokens that would otherwise cause performance issues.

\begin{table*}[ht]
\centering
\renewcommand{\arraystretch}{1} 
\resizebox{\textwidth}{!}{%
\begin{tabular}{lccccccc} \toprule
\multicolumn{1}{c}{\textbf{Model}} & \textbf{Fine-tuning Samples} & \textbf{AIME24} & \textbf{MATH-500} & \textbf{AMC23} & \textbf{Minerva} & \textbf{OlympiadBench} & \textbf{Avg.} \\ \midrule
\multicolumn{8}{l}{\cellcolor{white}\bf \textit{General Models}}       \\
Llama-3.1-70B-Instruct             &                             & 16.7            & 64.6              & 30.1           & 35.3             & 31.9                   & 35.7          \\
o1-preview                         &                          & 44.6            & 85.5              &        --        &           --       &                 --      &      --         \\ \midrule \midrule
\multicolumn{8}{l}{\cellcolor{white}\bf \textit{1.5B Models}}  \\
Still-3-1.5B-Preview               &       30,000                       & 32.5            & 84.4              & 66.7           & 29.0             & 45.4                   & 51.6          \\
DeepScaleR-1.5B-Preview            &      40,000                        & 43.1            & 87.8              & 73.6           & 30.2             & 50.0                   & 57.0          \\
Open-RS1                           &      18,615                        & 30.0            & 83.8              & 70.0           & 29.0             & 52.4                   & 53.0          \\
\rowcolor[HTML]{EFEFEF}\multicolumn{8}{c}{\textbf{1.5B Model Experiments}} \\
\rowcolor[HTML]{EFEFEF} DeepSeek-R1-Distill-Qwen-1.5B      &   \textbf{\textit{Base Model }}                         & 28.8            & 82.8              & 62.9           & 26.5              & 43.3                   & 48.9          \\
\rowcolor[HTML]{EFEFEF} GRPO      &   7000                         & 33.3            & 85.0               & 67.5           & 27.2               & 49.9                   & 52.6 (+3.7)          \\
\rowcolor[HTML]{EFEFEF} Clip-Cov~\citep{clip_cov}      &   7000                         & 33.3            & 85.5               & 70.0           & 29.0               & 50.0                   & 53.6 (+4.7)          \\
\rowcolor[HTML]{EFEFEF} \textbf{GRPO + Gaussian Reweight (ours)}     &   7000                         & 30.0            & 87.0              & 77.5           & 29.8             & 52.0                   & 55.3 (+6.4)          \\
\midrule
\midrule
\multicolumn{8}{l}{\cellcolor{white}\bf \textit{7B Models}}             \\
rStar-Math-7B                      &                              & 26.7            & 78.4              & 47.5           &       --           & 47.1                   &      --         \\
Eurus-2-7B-PRIME                   &                                  & 26.7            & 79.2              & 57.8           & 38.6             & 42.1                   & 48.9          \\
Qwen2.5-7B-SimpleRL                &                                  & 26.7            & 82.4              & 62.5           & 39.7             & 43.3                   & 50.9          \\
\rowcolor[HTML]{EFEFEF}\multicolumn{8}{c}{\textbf{7B Model Experiments}} \\
\rowcolor[HTML]{EFEFEF} Qwen-2.5-Math-7B-Instruct          &           \textbf{\textit{Base Model}}                    & 3.3            & 82.6              & 47.5           & 33.1             & 40.4                   & 41.4          \\
\rowcolor[HTML]{EFEFEF} GRPO        &   7000                         & 10.0             & 82.2              & 55.0           & 33.1             & 40.3                   & 44.1 (+2.7)            \\
\rowcolor[HTML]{EFEFEF} Clip-Cov~\citep{clip_cov} & 7000 & 10.0 (3/30) & 82.4 (412/500) & 57.5 (23/40) & 32.4 (88/272) & 41.3 (279/675) & 44.7 (+3.3) \\
\rowcolor[HTML]{EFEFEF} \textbf{GRPO + Gaussian Reweight (ours)}     &   7000                         & 13.3             & 82.8              & 62.5           & 32.0             & 42.7                   & 46.7 (+4.3)          \\
\bottomrule
\end{tabular}}
\caption{\textbf{Main Experimental Results.} This table presents zero-shot pass@1 performance across mathematical reasoning benchmarks. Values in parentheses indicate the improvement over the base model.}
\label{tab:main}
\end{table*}

\section{Experiments}
\subsection{Experimental Setup}

\paragraph{Models.}
We consider two different scales of models. For 1.5B models, we use DeepSeek-R1-Distill-Qwen-1.5B, which is Qwen2.5-Math-1.5B~\citep{qwen2.5math} fine-tuned on reasoning data from DeepSeek-R1~\citep{guo2025deepseekr1}. For 7B models, we use Qwen2.5-Math-7B-Instruct~\citep{qwen2.5math}.

\paragraph{Datasets.}
For training dataset, we use Open-RS~\citep{openrs}, which consists of 7k high quality math questions. For evaluation datasets, we use five broadly used math benchmarks. More details about the dataset information can be found in Appendix~\ref{app:datasets}.

\paragraph{Experimental Setup.}
Following~\citet{openrs}, we set the sampling temperature to 0.7 during training. The specific prompt template used for training is detailed in Appendix~\ref{app:prompt}, and comprehensive implementation details are provided in Appendix~\ref{app:implementation}.

\subsection{Results}
\paragraph{Accuracy Results.}
We present the experimental results in Table~\ref{tab:main}. Our covariance-weighted GRPO consistently outperforms vanilla GRPO across both model scales and all evaluation benchmarks. For the 1.5B DeepSeek-R1-Distill-Qwen model, our method achieves an average improvement of +2.7 points over vanilla GRPO, with particularly notable gains on AMC23~\cite{aime24} and Minerva~\cite{minerva}. Similarly, for the 7B Qwen2.5-Math model, our approach delivers a +2.4 point improvement on average, with the most substantial gain observed on AMC23 and OlympiadBench~\cite{he2024olympiadbench}. These consistent improvements across different model architectures and mathematical reasoning benchmarks demonstrate the effectiveness of our covariance-aware advantage reweighting in enhancing reasoning performance.




\paragraph{Policy Entropy Results.}
Using DeepSeek-R1-Distill-Qwen-1.5B, we plot the dynamics of entropy during training as introduced in Section~\ref{sec:motivation}. As shown in Figure~\ref{fig:entropy}, our proposed method maintains entropy at a reasonable level, effectively balancing exploration and exploitation, while vanilla GRPO exhibits significant entropy instability. To demonstrate that stable entropy correlates with better reasoning performance, we evaluate model performance at different training checkpoints on MATH-500~\cite{hendrycks2021measuring,lightman2023let} and OlympiadBench~\citep{he2024olympiadbench} to ensure more reliable statistical measurements, as shown in Table~\ref{tab:entropy_performance}.

The results confirm our hypothesis: vanilla GRPO's entropy instability leads to performance degradation (from 85.0\% to 79.8\% on MATH-500), while our method maintains consistent performance (86.2\%-87.0\%) across all checkpoints, validating the importance of controlling extreme covariance values.

\begin{table}
\centering
\footnotesize
\scalebox{0.86}{
\begin{tabular}{lccc}
\toprule
\textbf{Method} & \textbf{Training Step} & \textbf{MATH-500} & \textbf{OlympiadBench} \\
\midrule
\multirow{3}{*}{GRPO} 
& 100 & 85.0 & 49.9 \\
& 150 (entropy low) & 82.0 & 49.9 \\
& 200 (entropy high) & 79.8 & 47.8 \\
\midrule
\multirow{3}{*}{CW-GRPO} 
& 100 & 87.0 & 52.0 \\
& 150 (entropy low) & 86.2 & 53.9 \\
& 200 (entropy high) & 86.4 & 53.5 \\
\bottomrule
\end{tabular}
}
\caption{\textbf{Performance Comparison at Different Checkpoints.} Vanilla GRPO training exhibits entropy instability, resulting in degraded performance.}
\label{tab:entropy_performance}
\end{table}

\section{Conclusion}

We present a covariance-aware extension to GRPO that uses Gaussian kernel weighting to moderate extreme token-level updates, automatically improving the exploration-exploitation trade-off without additional hyperparameters. Experimental results across multiple reasoning benchmarks demonstrate consistent improvements over vanilla GRPO, validating our principled approach to enhancing reasoning performance through more balanced gradient updates.

\section*{Limitations}
Our experiments are conducted on models up to 7B parameters, and while the results demonstrate consistent improvements, further validation on larger-scale models would strengthen the evidence for the method's broad applicability. Additionally, our evaluation focuses primarily on mathematical reasoning tasks, which provide clear correctness criteria ideal for testing our approach. Exploring the method's effectiveness across more diverse tasks would offer broader insights into its general utility.

\section*{Acknowledgments}

We appreciate the reviewers for their insightful
comments and suggestions.
Qin Liu and Muhao Chen were supported by an Amazon Nova Trusted AI Prize, grants OAC 2531126 and ITE 2333736 from the United States National Science Foundation.

\bibliographystyle{acl_natbib}
\bibliography{custom}

\appendix
\section{Datasets Information}
\label{app:datasets}
We use Open-RS dataset as the training set, which is curated by~\citet{openrs}, totaling 7,000 samples: 3,000 from the Open-s1 dataset~\citep{openrs} (filtered mathematical problems from diverse sources like NuminaMATH~\citep{numina_math_datasets} and AIME), 3,000 from the Open-DeepScaleR~\citep{openrs} dataset (mathematics problems from AIME, AMC, and OmniMATH~\citep{Omni-MATH}), and 1,000 easier problems from the DeepScaleR~\citep{guo2025deepseekr1} dataset. Both models are trained on the Open-RS dataset.
For evaluation, we select five datasets: AIME24, MATH-500~\citep{hendrycks2021measuring,lightman2023let}, AMC23, Minerva~\citep{minerva} and OlympiadBench~\citep{he2024olympiadbench}. More information is presented in Table~\ref{tab:datasets_stats}.

\begin{table}[h]
\centering
\footnotesize
\resizebox{0.50\textwidth}{!}{
\begin{tabular}{lcc}
\toprule
\textbf{Name} & \textbf{Huggingface Path} & \textbf{Size} \\
\midrule
AIME24         &  HuggingFaceH4/aime\_2024      &  30   \\
AMC23      &   knoveleng/AMC-23     &   40  \\
MATH-500 & HuggingFaceH4/MATH-500 & 500 \\
Minerva & knoveleng/Minerva-Math & 272 \\
OlympiadBench & knoveleng/OlympiadBench & 675 \\
\bottomrule
\end{tabular}
}
\caption{\textbf{Datasets Information.}}
\label{tab:datasets_stats}
\end{table}

\section{Training Prompt}
\label{app:prompt}
The prompt used during training is presented in Figure~\ref{fig:prompt}, in which we instruct the model to use English only, as we have observed some language mixture issues.
\begin{figure}[h]
\centering
\begin{tcolorbox}[fonttitle=\bfseries,title=Prompt Used for Training, size = normal]
A conversation between User and Assistant. The user asks a question, and the Assistant solves it. The assistant first thinks about the reasoning process in the mind and then provides the user with the answer, and put your final answer within boxed{{}} . The reasoning process and answer are enclosed within <think> </think> and <answer> </answer> tags, respectively, i.e., <think> reasoning process here </think> <answer> answer here </answer>. Note that respond by English, NOT use other languages.
\end{tcolorbox}
\caption{prompt}
\label{fig:prompt}
\end{figure}

\section{Implementation Details}
\label{app:implementation}
We conduct all training and evaluation using four NVIDIA H200 GPUs. Our reward function combines accuracy and format metrics. For accuracy, we compare the parsed model output against the ground truth using a verification function implemented in \LaTeX. The function assigns a reward of 1.0 for exact matches and 0.0 otherwise. For format compliance, we award a reward of 1.0 when the output contains properly matched \texttt{<think>} tags. 

Training is performed using the HuggingFace \texttt{trl} framework\footnote{\url{https://github.com/huggingface/trl}}, while evaluation utilizes the HuggingFace \texttt{lighteval} framework\footnote{\url{https://github.com/huggingface/lighteval}}. The specific hyperparameters used in our experiments are detailed in Table~\ref{tab:hyperparameters}.
\begin{table}[h]
\centering
\footnotesize
\begin{tabular}{cc}
\toprule
\textbf{Parameter} & \textbf{Value} \\
\midrule
Learning Rate & $1.0 \times 10^{-6}$ \\
Batch Size & 12 \\
Gradient Accumulation Steps & 4 \\
Training Steps & 100 \\
Warmup Ratio & 0.1 \\
Max Prompt Length & 512 \\
Max Completion Length & 4096 \\
Temperature & 0.7 \\
Number of Generations & 12 \\
\bottomrule
\end{tabular}
\caption{\textbf{Hyperparameters Configuration for Experiments.}}
\label{tab:hyperparameters}
\end{table}

\section{Related Work}
\paragraph{Test-time Scaling for Reasoning Tasks.}
Test-time scaling has emerged as a promising paradigm for improving LLM performance by allocating additional computational resources during inference. \citep{snell2024scalingllmtesttimecompute} demonstrated that scaling test-time compute optimally can be more effective than scaling model parameters, showing over 4× efficiency gains through compute-optimal strategies. \citep{muennighoff2025s1} introduced a simplified approach using "budget forcing" to control inference compute by appending "Wait" tokens, achieving strong reasoning with minimal training data. \citep{zhao2025genprmscalingtesttimecompute} advanced the field with GenPRM, a generative process reward model that scales test-time compute through explicit Chain-of-Thought reasoning. \citep{setlur2024rewardingprogressscalingautomated} proposed that effective process rewards should measure progress by evaluating likelihood changes before and after each reasoning step.

\paragraph{Reinforcement Learning for Reasoning Tasks.}
Reinforcement Learning with Verifiable Rewards (RLVR) has rapidly become the dominant route for eliciting step-by-step reasoning in LLMs.  \citet{shao2024deepseekmath} first showed that Group Relative Policy Optimization (GRPO) can improve while dispensing with a value network, and the recipe was later scaled in DeepSeek-R1~\citep{deepseekai2025deepseekr1incentivizingreasoningcapability}.  Subsequent work diagnoses exploration bottlenecks: unlikeliness reward boosts low-probability but correct trajectories~\citep{he2025rewarding}, whereas covariance-based clipping traces early saturation to entropy collapse~\citep{cui2025entropy,hao2025rethinking}.  Efficiency studies reveal that even a \emph{single} worked example can unlock large gains through 1-shot RLVR~\citep{wang2025oneshot}.

\end{document}